Original Paper

# Association Between Neighborhood Factors and Adult Obesity in Shelby County, Tennessee: Geospatial Machine Learning Approach

Whitney S Brakefield[1,2], PhD; Olufunto A Olusanya[2], MPH, MD, PhD; Arash Shaban-Nejad[2], MPH, PhD

[1]Bredesen Center for Data Science and Engineering, University of Tennessee, Knoxville, TN, United States
[2]Center for Biomedical Informatics, Department of Pediatrics, College of Medicine, University of Tennessee Health Science Center, Memphis, TN, United States

**Corresponding Author:**
Arash Shaban-Nejad, MPH, PhD
Center for Biomedical Informatics
Department of Pediatrics
College of Medicine, University of Tennessee Health Science Center
50 N Dunlap Street, 491R
Memphis, TN, 38103
United States
Phone: 1 901 287 5836
Email: ashabann@uthsc.edu

## Abstract

**Background:** Obesity is a global epidemic causing at least 2.8 million deaths per year. This complex disease is associated with significant socioeconomic burden, reduced work productivity, unemployment, and other social determinants of health (SDOH) disparities.

**Objective:** The objective of this study was to investigate the effects of SDOH on obesity prevalence among adults in Shelby County, Tennessee, the United States, using a geospatial machine learning approach.

**Methods:** Obesity prevalence was obtained from the publicly available 500 Cities database of Centers for Disease Control and Prevention, and SDOH indicators were extracted from the US census and the US Department of Agriculture. We examined the geographic distributions of obesity prevalence patterns, using Getis-Ord Gi* statistics and calibrated multiple models to study the association between SDOH and adult obesity. Unsupervised machine learning was used to conduct grouping analysis to investigate the distribution of obesity prevalence and associated SDOH indicators.

**Results:** Results depicted a high percentage of neighborhoods experiencing high adult obesity prevalence within Shelby County. In the census tract, the median household income, as well as the percentage of individuals who were Black, home renters, living below the poverty level, 55 years or older, unmarried, and uninsured, had a significant association with adult obesity prevalence. The grouping analysis revealed disparities in obesity prevalence among disadvantaged neighborhoods.

**Conclusions:** More research is needed to examine links between geographical location, SDOH, and chronic diseases. The findings of this study, which depict a significantly higher prevalence of obesity within disadvantaged neighborhoods, and other geospatial information can be leveraged to offer valuable insights, informing health decision-making and interventions that mitigate risk factors of increasing obesity prevalence.

(*JMIR Public Health Surveill 2022;8(8):e37039*)   doi: 10.2196/37039

**KEYWORDS**

obesity; obesity surveillance; disease surveillance; machine learning; geographic information systems; social determinants of health; SDOH; disparities

XSL•FO
RenderX

## Introduction

Obesity is a global epidemic with increasing prevalence from 3% to 11% among men and 6% to 15% among women within the past 40 years [1]. Obesity is responsible for at least 2.8 million deaths per year [2] and is defined by the Centers for Disease Control and Prevention (CDC) as "weight that is higher than what is considered healthy for a given height" with a BMI of 30 kg/m$^2$ or higher [3]. It is a noncommunicable disease (NCD) that could have a profound, lifelong adverse impact on individuals' overall life expectancy, quality of life, and other clinical outcomes. Moreover, obesity increases susceptibility to developing other NCDs such as diabetes mellitus, hypertension, heart disease, myocardial infarction, stroke, fatty liver disease, and cancers. According to the CDC, obesity is associated with the top leading causes of death in the United States. With over 42% of individuals living with obesity, there is a significant US $147 billion financial burden placed on the United States [4].

Although genetic and behavioral factors increase susceptibility, studies have shown that social determinants of health (SDOH) risk factors adversely affect health outcomes and are major contributing factors to the increasing occurrence of obesity and other NCDs [5-9]. Evidence suggests that the pattern of distribution for societal resources and socioeconomic status are correlated with the quality-of-life attributes as well as physical and psychosocial characteristics [10]. SDOH indicators including education attainment, financial security, health literacy, access to healthy food, poverty level, employment conditions, and health care access are determined to be the most significant predictors of an individual's health status. Moreover, SDOH indicators are perceived to be among major driving forces behind systematic social inequalities [11]. As a result, certain susceptible populations are more likely to be affected by obesity-associated SDOH stressors than other groups and populations [12]. The ongoing global pandemic caused by COVID-19 has further worsened SDOH burdens, since individuals diagnosed with preexisting conditions such as obesity have been disproportionately affected by COVID-19 morbidity and mortality [13]. Tailored and effective obesity prevention interventions should be implemented within the context of sociocultural, socioeconomical, environmental, psychosocial, and demographic indicators that influence population health.

There is a dearth of studies that have leveraged geospatial intelligence to examine SDOH indicators associated with obesity. In this study, we examined the geographical variations and prevalence patterns of obesity in Shelby County in the United States, using Getis-Ord Gi* statistics and calibrated multiple models to study the association between SDOH and adult obesity. We also adopted unsupervised machine learning to conduct grouping analysis and investigate the distribution of obesity prevalence and the associated SDOH indicators. In addition to facilitating the surveillance of obesity and other NCDs within Shelby County, our findings could inform innovative health strategies to tackle SDOH disparities and other adverse influences on health outcomes.

## Methods

### Data Source

In this study, data from well-known, publicly available multidimensional sources were merged at the census tract level. We used CDC 500 Cities data (2019) [14], which represents city-level data originating from 500 largest US cities, to determine obesity prevalence. The CDC 500 Cities data were merged with SDOH data extracted from the American Community Survey and the US Department of Agriculture (2018-2020) estimates [15,16]. Table 1 shows the summary statistics for variables included in the study.

**Table 1.** Summary statistics for obesity and related risk factors in census tracts of Shelby County, Tennessee.

| Variables | Operationalization | Source | Values, mean (SD) |
|---|---|---|---|
| Obesity | Model-based estimate for the crude prevalence of obesity among adults aged ≥18 years, 2018 | CDC[a] | 35.77 (7.84) |
| Low access to supermarket | Count of housing units without a vehicle and greater than half a mile from supermarket in the census tract | USDA[b] | 102.54 (108.37) |
| Black population | Percentage of the Black or African American population living in the census tract | US census | 58.02 (17.31) |
| Poverty | Percentage of the population living below the federal poverty line in the census tract | USDA | 24.89 (17.35) |
| Unemployment | Percentage of the unemployed population living in the census tract | US census | 4.32 (3.04) |
| High school diploma | Percentage of the population aged ≥25 years without a high school diploma in the census tract | US census | 9.33 (6.59) |
| Renters | Percentage of the population renting their homes | US census | 18.87 (11.85) |
| Average household size | Average household size in a census tract | US census | 2.57 (0.52) |
| Median household income | Median household income in a census tract (US $) | US census | 53,746 (29,335) |
| Female head of the household | Percentage of the households with a female head in a census tract | US census | 7.75 (4.23) |
| Uninsured | Model-based estimate for the crude prevalence of uninsured adults aged ≥18 years, 2018 | CDC | 18.84 (7.16) |
| Lack of physical activity | Model-based estimate for the crude prevalence of lack of physical activity among adults aged ≥18 years, 2018 | CDC | 32.88 (10.52) |
| Aged 55 years and older | Percentage of the population aged ≥55 years in a census tract | US census | 21.89 (7.81) |
| Single | Percentage of the population who are single in a census tract | US census | 13.70 (8.62) |

[a]CDC: Centers for Disease Control and Prevention.
[b]USDA: The United States Department of Agriculture.

## Obesity Clusters

We explored the geospatial clustering and hot spots of adult obesity prevalence in Shelby County. We conducted this analysis by using Getis-Ord Gi* statistics with first order queen contiguity and applied the false discovery rate correction parameter to account for multiple testing and spatial dependence.

## Regression Modeling

### Data Wrangling

To prepare the data set for predictive modeling, we scaled our features such that columns had a mean of 0 and a SD of 1 [17]. Relative scales have been shown to reduce heterogeneity and allow for variable comparison [18].

### Model Selection

The predictor variables that were considered were the 13-census, tract-level risk factor variables, and the outcome variable was the adult obesity prevalence in the census tract (Table 1). We used the "forward and backward" stepwise regression to depict a subset of the variables and Akaike's information criterion (AIC) as the metric [19,20]. Variance inflation factor (VIF) was applied to assess redundancy between predictor variables to prevent multicollinearity. VIF factors that exceeded 10 were removed [21]. Predictor variable values that were not significant ($P<.05$) were removed.

### Models

In this study, we applied multiple modeling techniques. Ordinary least squares (OLS) regression modeling was amongst these techniques, represented by the following equation:

$$Y = X\beta + \varepsilon \quad (1)$$

Equation 1 shows the regression model in matrix notation, where Y is an n×1 vector of n observations on the dependent variable; X is an n×q design matrix of n observations on q explanatory variables (first column in X matrix will consist of a vector of n ones for the intercept); β is a q×1 vector of regression coefficients; and ε represents an n×1 vector of random error terms (independently and identically distributed). To assess and compare the performance of the models, we used adjusted $R^2$ and AIC. To assess the heteroskedasticity of random error terms, we used the Koenker-Bassett test. To assess the normality of error distribution, the Jarque-Bera test was applied. We assessed the multicollinearity of the entire model using the condition number. To examine the independence of the terms Robust Lagrange Multiplier (error) and Robust Lagrange Multiplier (lag) methods were applied. First, order queen contiguity weights were constructed for spatial testing. Queen contiguity was chosen because areas sharing all boundaries and vertices are considered as neighbors, which yields more neighbors per area than the rook contiguity. If dependence was found among the terms, we incorporated the terms that accounted for autocorrelation in the model. Thus, we applied spatial autoregressive models: spatial lag or spatial error model (SEM)

[22]. The spatial lag model includes a spatially lagged dependent variable and is represented by equation 2:

$$Y = X\beta + \rho WY + \varepsilon \quad (2)$$

In equation 2, Y is an n×1 vector of n observations on the dependent variable; ρ is a scalar spatial lag parameter; WY is the spatially lagged dependent variable for an n×n weights matrix W; X is an n×q design matrix of n observations on q explanatory variables; β is an q×1 vector of regression coefficients; and ε represents an n×1 vector of error terms.

The spatial error model includes a spatial autoregressive error term and is represented by equation 3:

$$Y = X\beta + u,\ u = \lambda Wu + \varepsilon \quad (3)$$

In equation 3, Y is an n×1 vector of n observations on the dependent variable; X is an n×q design matrix of n observations on q explanatory variables; β is an q×1 vector of regression coefficients; λ is a scalar spatial error parameter; W represents the n×n spatial weights matrix; u represents an n×1 vector of error terms; Wu denotes a spatially lagged error term; and the represents an n×1 vector of error terms. OLS regression and spatial autoregressive models will be assessed and compared to depict the optimal performance.

### Grouping Analysis

In order to understand the dependent variable and significantly associated SDOH across the region, we used the hierarchical clustering unsupervised machine learning algorithm [23-25] in the "stats" package embedded in R software (version 4.0.3; RStudio, PBC) [25] to conduct an exploratory grouping analysis. Ward's Method was used to minimize the increase in the error sum of squares [26].

### Lack of Physical Activity, Obesity, and SDOH

We explored the geographical distribution of lack of physical activity, obesity, and the top four features significantly associated with obesity in Shelby County.

### Visualization and Tools

ArcGIS Pro software (version 2.9.0; Esri) was used to produce spatial distributions to investigate patterns (ie, spatial clustering). R Studio (version 4.0.3; RStudio, PBC) and GeoDa software (version 1.16.0.12; Luc Anselin) were used for statistical analyses.

## Results

### Obesity Clusters

Figure 1 reflects adult obesity prevalence geospatial distribution and adult obesity prevalence significant clusters in the study region.

Figure 1A shows a high percentage of the population in the central and southwestern regions diagnosed with adult obesity, and Figure 1B shows that the central western region is also a significant hot spot for adult obesity. Conversely, significant cold spots are clustered along the eastern region of the Shelby County.

**Figure 1.** (A) represents geospatial distribution of adult obesity prevalence in Shelby County; (B) represents significant hot and cold spots of adult obesity prevalence in Shelby County.

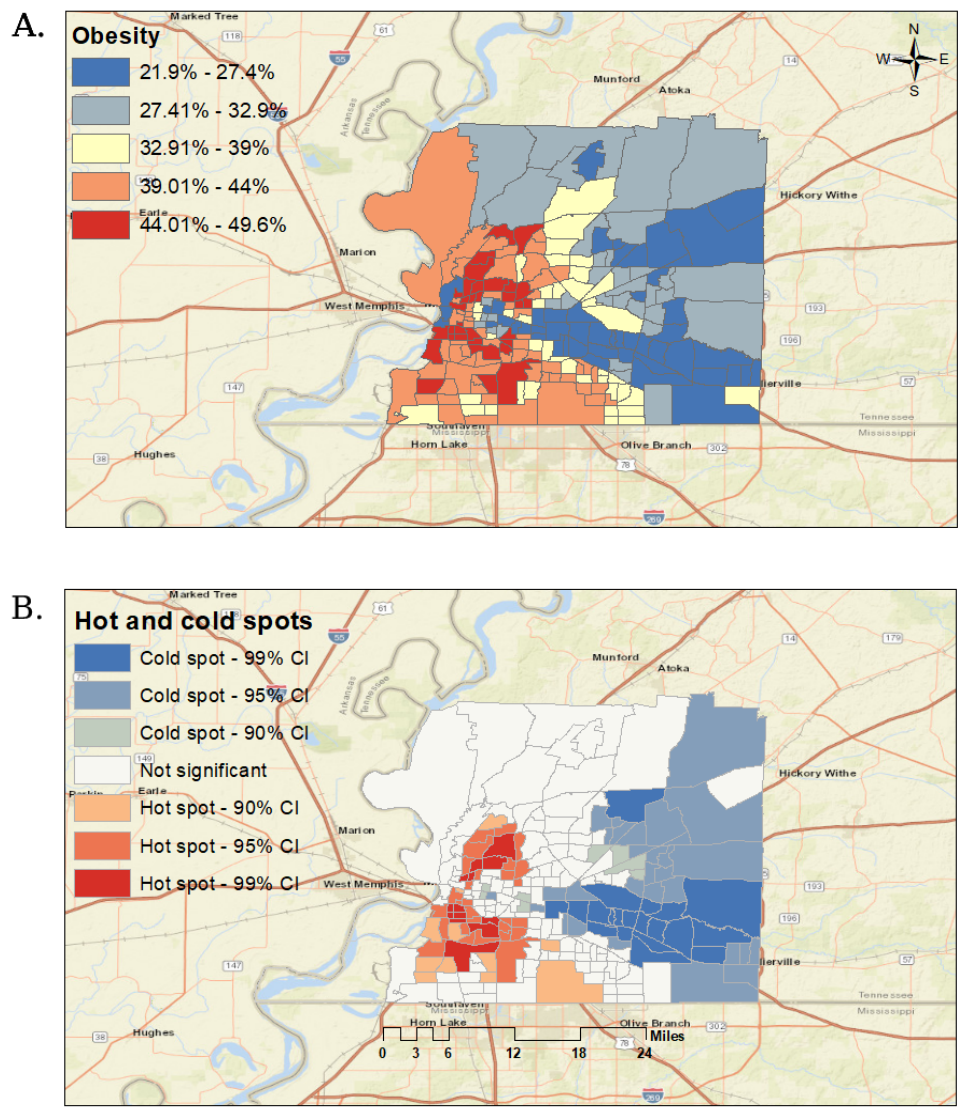

## Model Selection

After conducting the analytical modeling techniques in the "Regression Modeling" section, the percentage of population that lacks physical activity was removed during the VIF assessment (VIF=46.7), and the percentage of population with a female head of the household and the percentage of the population aged 25 years and older without high school education were removed during the AIC process (they were also found to be nonsignificant after conducting further experimental analysis). In addition, the average household size and households with low access to supermarkets were not significantly associated with obesity. However, there were 8 variables from Table 1 that were significantly associated with obesity prevalence: median household income, percentage of the Black population, poverty level, percentage of the uninsured population, percentage of the population aged 55 years or older, percentage of the population who are single, percentage of the unemployed population, and percentage of home renters. The significant variables each had a VIF ≤10.0.

## The Final OLS Model Results

The final OLS regression model results are shown in Table 2, which displays the predictor variables that best describe the model. The adjusted $R^2$ was 0.963, indicating that 96% of the variation in the outcome variable was explained by the predictors with an AIC of –88.34. There was a multicollinearity condition number of 6.99, which is less than 20, thus not suspected of multicollinearity. The Jarque-Bera test had a $P$ value <.001. Koenker-Bassett test had a $P$ value of .17, indicating the presence of constant variance in error terms. The $P$ value ($F$ statistics) less than .05 was deemed as significant or meaningful.

**Table 2.** Ordinary least squares regression results.

| Variable | Coefficient |
| --- | --- |
| Constant | –0.000 |
| Median household income | –0.046[a] |
| Poverty | 0.184[b] |
| Renters | –0.134[b] |
| Aged 55 years and older | 0.043[a] |
| Single | 0.091[c] |
| Uninsured | 0.445[b] |
| Unemployment | 0.042[a] |
| Black population | 0.438[b] |

[a]$P<.05$.

[b]$P<.001$.

[c]$P<.01$.

However, Robust Lagrange Multiplier (error) had a test value of 10.72 ($P=.001$), which was significantly higher than Robust Lagrange Multiplier (lag) with a test value of 8.449 ($P=.003$). OLS model results are not reliable due to significant spatial dependency. A spatial error term will be incorporated into the model.

**Spatial Error Model**

Table 3 shows the SEM results. In the model, the percentage of the Black population, the percentage of the population below poverty rate, the percentage of the population who are single, the percentage of uninsured population, and the percentage of the population aged 55 years or older are positively associated with obesity, showing an increase in obesity. On the other hand, the median household income and the percentage of home renters are negatively associated with obesity, showing a decrease in obesity.

Since our variables are measured on the same scale, we were able to compare the strength of the effect of each predictor variable on obesity prevalence. We found that the percentage of uninsured population, the percentage of the Black population, the percentage of the population below poverty level, and the percentage of home renters were the most important variables when predicting obesity prevalence in Shelby County.

**Table 3.** Spatial Error Model results.

| Variable | Coefficient |
| --- | --- |
| Constant | –0.001 |
| Lambda | 0.488[a] |
| Median household income | –0.056[a] |
| Renters | –0.106[a] |
| Poverty | 0.146[a] |
| Aged 55 years or older | 0.051[b] |
| Single | 0.066[c] |
| Uninsured | 0.466[a] |
| Unemployment | 0.027 |
| Black population | 0.423[a] |

[a]$P<.001$.

[b]$P<.01$.

[c]$P<.05$.

## Overall Model Performance Comparison

After calibrating both models, we found that SEM outperformed the OLS model. Table 4 shows that the $R^2$ value improved to 0.968 after incorporating the error term in the model, and the AIC improved to –108.09, indicating a better model fit.

**Table 4.** Model performance.

| Model | Adjusted $R^2$ | Akaike's information criterion |
| --- | --- | --- |
| Ordinary least squares | 0.963 | –88.34 |
| Spatial error model | 0.968 | –108.09 |

## Grouping Analysis

Our grouping analysis divided the study area into 5 distinct groups across the Shelby region, based on the top four features that were significantly associated with obesity (Figure 2).

**Figure 2.** Grouping analyses results.

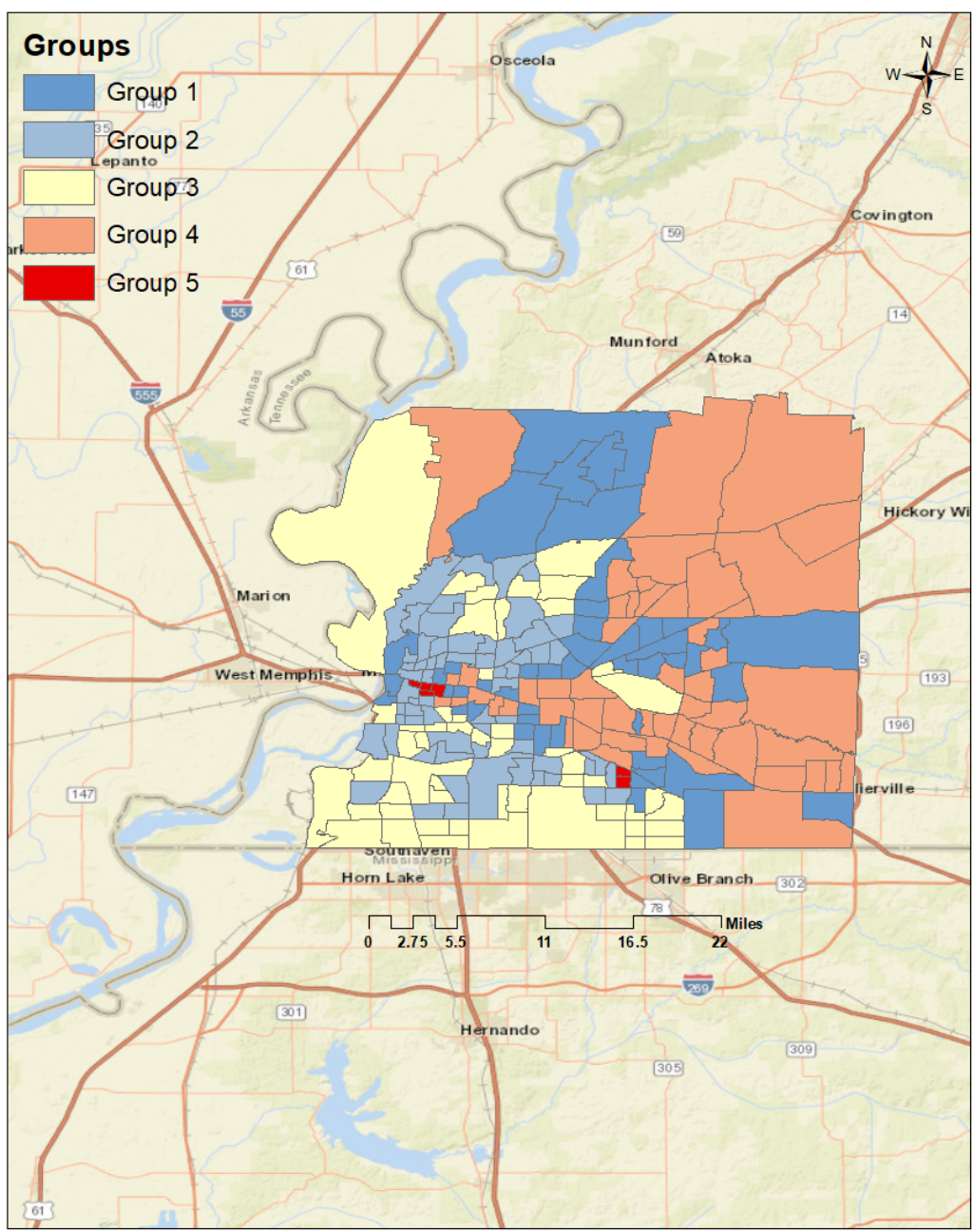

Group 1 spans the fourth largest area of the region (47 census tracts) and was quantified as being below average in obesity prevalence, percentage of the Black population, percentage of the population with an income below the poverty level, and percentage of the uninsured population; however, this group is around average in the percentage of renters.

Group 2 is the largest area in the region, comprising of 62 census tracts. This region is far above average in obesity prevalence, percentage of renters, percentage of the Black population, percentage of the population with an income below the poverty level, and percentage of the uninsured population.

Group 3 comprises of 52 census tracts. This region is above average in obesity prevalence, percentage of renters and percentage of the uninsured population, and it is far above average in percentage of the Black population; however, this group is around average in percentage of the population with an income below the poverty level and below average in percentage of renters.

Group 4 comprises of 52 census tracts and is quantified as being far below average in obesity prevalence, percentage of the Black population, percentage of the population with an income below the poverty level, percentage of renters, and percentage of the uninsured population.

Group 5 spans the smallest area of the region (6 census tracts) and is characterized as being average in obesity prevalence and percentage of the uninsured population; however, this group is far above average in percentage of the Black population, percentage of the population with an income below the poverty level, and percentage of renters.

## Lack of Physical Activity, Obesity, and SDOH

Even though lack of physical activity was removed during the "model selection" process due to multicollinearity, we examined the Spearman rank correlation coefficient (Table 5), the geospatial distribution of obesity (Figure 1A), and lack of physical activity (Figure 3), as well as the geospatial patterns among the top four obesity-associated features and lack of physical activity (Figure 3). The Spearman rank coefficient shows a strong positive relationship between lack of physical activity and obesity. Figure 1A shows a high prevalence of obesity clusters in the central and southwestern regions of Shelby County, consistent with the top four obesity-associated features and the lack of physical activity geospatial pattern.

In addition, Table 5 shows a strong positive relationship between lack of physical activity and the top four features associated with obesity. Geographically, we found that the central and southwestern regions of Shelby County consisted of a high percentage of population who are below the poverty rate, Black, and uninsured, and the percentage of the population who lack physical activity was consistent with these geospatial patterns. On the other hand, the eastern region of Shelby County showed a consistent pattern among the low percentage of the population below poverty rate, percentage of the Black population, percentage of renters, and percentage of the uninsured population, and consisted of clustered census tracts that contained a low percentage of the population who lack physical activity.

**Table 5.** Spearman rank coefficients to assess the relationship between lack of physical activity and obesity and the top four obesity-associated features in Shelby County census tracts.

| Variables | Spearman rank coefficient |
| --- | --- |
| Obesity | 0.96[a] |
| Uninsured population | 0.95[a] |
| Black population | 0.76[a] |
| Renters | 0.43[a] |
| Poverty | 0.86[a] |

[a]$P<.001$.

**Figure 3.** Assessment of lack of physical activity and the top four features associated with obesity.

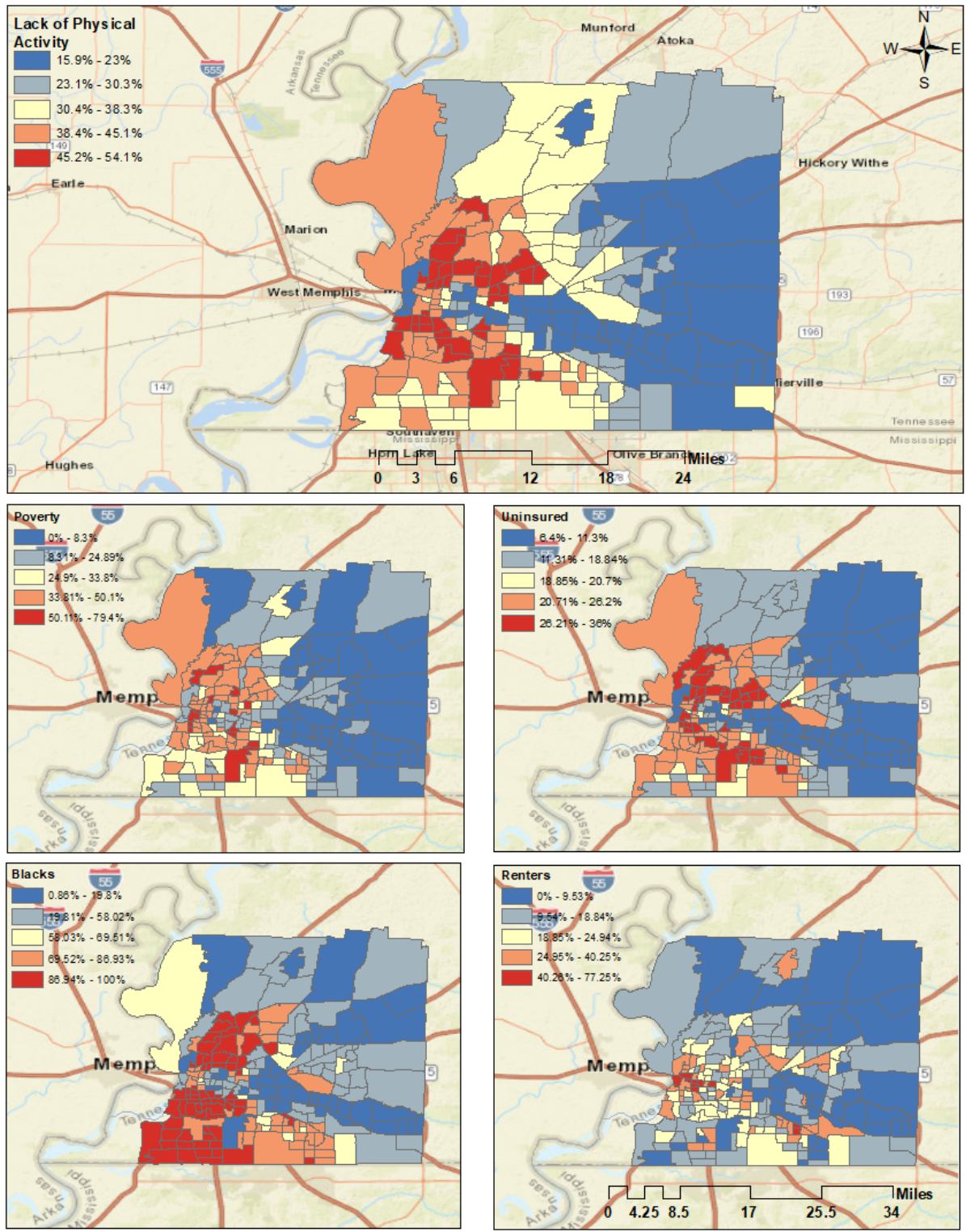

## Discussion

Obesity is a serious health condition that is associated with several comorbidities (eg, heart diseases, cancers, and diabetes) that are leading causes of death in the United States. SDOH factors such as the community, home, school, and workplace setting can impact physical activity and access to affordable healthy food. Some communities are more impacted, as evidenced by the disproportionality of adult obesity rates, compared to other populations [27,28]. Although a few studies have leveraged geospatial analysis in the United States to explore the relationship between neighborhood factors and obesity, this study was a critical step in understanding and effectively addressing chronic diseases. Using Getis-Ord Gi* statistics and unsupervised machine learning, this study examined how SDOH characteristics influenced obesity prevalence among adults 18 years or older in Shelby County. In a study by Cohen et al [29] in 2017, obesity rates were modeled against urban-rural geographic status, using the Behavioral Risk Factor Surveillance System. Moreover, our findings, showing associations between SDOH indicators (eg, race, income level, and poverty rate) and obesity, are consistent with the findings of other studies [29-31]. Our study also found that in the eastern region of Shelby County, where the percentage of home renters was low (Figure 3), the obesity rate was also low (Figure 1A). Thus, a population's rental status could play a role in the obesity prevalence. However, contrary to other studies [32-34], our study found that lack of educational attainment was not significantly associated with an increase in obesity prevalence. Given some of the SDOH risk factors that have been identified (eg, percentage of the population below poverty rate, low median household income, percentage of renters, Black population, and the uninsured population), as well as the high obesity prevalence depicted among socially disadvantaged groups within Shelby County, our study proposes that the effective planning and implementation of intervention strategies to address obesity are informed by SDOH surveillance. Notably, our model calibration results indicate that SEM outperformed the OLS model.

Unlike multiple studies [5-12,27-34] that have examined obesity and SDOH, we provided an analysis to assess adult obesity and SDOH at the census tract level in Shelby County, Tennessee. Admittedly, some limitations should be considered with our findings. First, cross-sectional studies such as ours are unable to detect causal relations between predictor and outcome variables nor are they able to qualitatively examine sociocontextual influences. Another limitation is that when aggregating data such as SDOH and analyzing at a specific level of granularity, a change in units could alter the findings (modifiable areal unit problem). In addition, our study may not be generalizable to the whole Tennessee state and the United States. In the future, we will conduct comparative studies to assess the generalizability of our results and include additional SDOH indicators (eg, proximity to green spaces, crime, and transportation) and social and community contexts (eg, social cohesion) to expand our study. In addition, 500 Cities only provide data for 219 of 221 census tracts in Shelby County, which could pose a problem during the integration process; we removed the missing census tracts (ie, 980200 and 980300) from other integrated data sets for parallelism. Another limitation is that CDC 500 Cities data relies on self-reported surveys that have not been continuously scrutinized for potential social desirability bias and measurement bias. However, this data set offers access to validated, regionally representative data. Despite these limitations, our study depicts that SDOH and environmental characteristics at a granular level are major risk factors for obesity in Shelby County.

Finally, results from this study will be incorporated into the analytics layer of our Urban Public Health Observatory knowledge-based surveillance platform [35,36] and Personal Health Libraries [37]. These platforms could facilitate the collection of public health evidence as well as surveillance data that will facilitate precision population health [38] to provide immediate insights to inform decision-making at multiple levels of authority, including among health officials, patients, physicians, and caregivers.

## Conclusion

Previous studies have found associations between sociogeographical determinants and health outcomes [39-42]. Likewise, our study found that a high percentage of disadvantaged neighborhoods within the Shelby region had significantly higher obesity prevalence and SDOH risk factors. Accordingly, policies should be implemented that are socioculturally adaptable and tailored toward vulnerable communities and can address SDOH disparities and are geared to tackle underlying determinants of the obesity epidemic.

## Conflicts of Interest

None declared.

## Abbreviations

**AIC:** Akaike's information criterion
**CDC:** Centers for Disease Control and Prevention
**NCD:** noncommunicable disease
**OLS:** ordinary least squares
**SDOH:** social determinants of health
**SEM:** spatial error model
**VIF:** variance inflation factor